\documentclass[conference,9pt]{IEEEtran}
\IEEEoverridecommandlockouts

\usepackage{cite}
\usepackage{amsmath,amssymb,amsfonts}
\usepackage{algorithmic}
\usepackage{graphicx}
\usepackage{textcomp}
\usepackage{xcolor}
\usepackage{array} 
\usepackage{multirow} 
\usepackage{diagbox}
\usepackage{booktabs}
\usepackage{url}

\def\BibTeX{{\rm B\kern-.05em{\sc i\kern-.025em b}\kern-.08em
    T\kern-.1667em\lower.7ex\hbox{E}\kern-.125emX}}
\begin{document}

\title{Do Multimodal Language Models Really Understand Direction? A Benchmark for Compass Direction Reasoning\\
}

\author{
    \IEEEauthorblockN{Hang Yin,  Zhifeng Lin, 3\textsuperscript{rd} Xin Liu,  Bin Sun,  Kan Li}
    \IEEEauthorblockA{
        \textit{Department of Computer Science and technology} \\
        \textit{Beijing Institute of Technology} \\
        Beijing, China \\
        \{yh, linzf, xinliu, binsun, likan\}@bit.edu.cn
    }
}

\maketitle

\begin{abstract}

Direction reasoning is essential for intelligent systems to understand the real world. While existing work focuses primarily on spatial reasoning, compass direction reasoning remains underexplored. To address this, we propose the Compass Direction Reasoning (CDR) benchmark, designed to evaluate the direction reasoning capabilities of multimodal language models (MLMs). CDR includes three types images to test spatial (up, down, left, right) and compass (north, south, east, west) directions. Our evaluation reveals that most MLMs struggle with direction reasoning, often performing at random guessing levels. Experiments show that training directly with CDR data yields limited improvements, as it requires an understanding of real-world physical rules. We explore the impact of mixdata and CoT fine-tuning methods, which significantly enhance MLM performance in compass direction reasoning by incorporating diverse data and step-by-step reasoning, improving the model's ability to understand direction relationships.
\end{abstract}


\section{Introduction}

The exploration of models' ability to understand the physical world has garnered significant attention~\cite{ha2018world, hao2023reasoning, liu2024world, xu2024penetrative}. Among these capabilities, direction reasoning stands out as a crucial cognitive skill for decision-making and navigation in the real world. This ability enables individuals to infer and understand the relative positions and orientations of objects in space, forming a foundation for effective communication and interaction with the environment. For intelligent systems, such as autonomous vehicles, robotics, and augmented reality, accurate direction reasoning is critical for tasks like path planning, object localization, and spatial awareness~\cite{DBLP:conf/aaai/ShiZL22,DBLP:journals/tmlr/YamadaBLKY24, mirzaee2021spartqa,DBLP:journals/corr/abs-2310-03249,DBLP:conf/aaai/LiH024a}. With the advancement of large language models (LLMs), they have demonstrated remarkable capabilities in object detection, image captioning, and image-based dialogue\cite{achiam2023gpt,radford2021learning,hu2022scaling, DBLP:conf/nips/LiuLWL23a,DBLP:journals/apin/YinLLSL24}. However, their ability to reason about directions, specifically compass direction reasoning that follows real-world direction rules, remains relatively unexplored.

Neglecting compass reasoning limits model effectiveness in applications like navigation, geographic positioning, and large-scale environment interactions~\cite{canny1988complexity,lavalle2006planning,lopes2000intelligent}. Autonomous systems, for instance, must use compass-based directions, not just relative spatial ones. Without accurate compass reasoning, they risk failure in tasks like path planning or location-based services, where consistent real-world orientation is crucial. Inconsistencies between spatial and compass reasoning can also disrupt communication in human-AI interactions, where instructions often rely on absolute directions (e.g., “head north”).

\setlength{\extrarowheight}{0.1pt} 
\begin{table*}[h]
    \centering
    \caption{Data examples in CDR benchmark}
    \label{tab:prompts}
    \newcolumntype{C}[1]{>{\centering\arraybackslash}m{#1}} 
    \newcolumntype{L}[1]{>{\raggedright\arraybackslash}m{#1}} 
    \begin{tabular}{C{0.4\columnwidth}|C{0.2\columnwidth}|L{1\columnwidth}|L{0.18\columnwidth}}
        \hline
        \textbf{Type} & \textbf{Image} & \centering \textbf{ Question} & \textbf{Answer} \\ \hline
        Object Classification & \multirow{4}{*}[-2.5ex]{\includegraphics[width=0.08\textwidth, clip, trim=72.5mm 40mm 195mm 110mm]{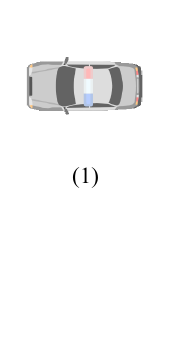}} & Please select the most appropriate option. Which of the following options best fits the \textcolor{red}{category} of the object in the picture? \newline  A. finger B. arrow  C. car D. plane  & C. car \\ \cline{1-1} \cline{3-4}
        Spatial Direction Reasoning & & Please select the most appropriate option. What \textcolor{red}{direction} does the car in the picture point to?  \newline  A. down B. left C. lower\_left D. lower\_right E. right F. up G. upper\_left H. upper\_right & F. up \\ \cline{1-1} \cline{3-4}
        Compass Direction Reasoning & & Please select the most appropriate option. What \textcolor{red}{direction} does the car in the picture point to in a plane coordinate system \textcolor{red}{with the upper part being West}?  \newline  A. East B. West C. South D. North E. Northeast F. Northwest G. Southeast H. Southwest  & B. West \\ \cline{1-1} \cline{3-4}
        \hline
        Relative Spatial Reasoning & \multirow{2}{*}[2.3ex]{\includegraphics[width=0.26\textwidth, clip, trim=185mm 114.5mm 40mm 63mm]{icon.pdf}} & Please select the most appropriate option. In a Cartesian coordinate system established on the plane, \textcolor{red}{with aircraft as the origin}, then in which direction is \textcolor{red}{flower relative to origin}?    \newline A. down B. left C. lower\_left D. lower\_right E. right F. up G. upper\_left H. upper\_right & G. upper\_left \\ \cline{1-1} \cline{3-4}
        Relative Compass Reasoning &  & Please select the most appropriate option. In a Cartesian coordinate system established on the plane, \textcolor{red}{with aircraft as the origin}. \textcolor{red}{If the nose of the aircraft is pointing South}, then in which direction is flower relative to origin?  \newline  A. East B. West C. South D. North E. Northeast F. Northwest G. Southeast H. Southwest & G. Southeast\\ \hline
        
        Relative Compass Reasoning (letter) & \raisebox{-0.3\height}{\includegraphics[width=0.09\textwidth, clip, trim=72.5mm 116mm 195mm 61.6mm]{icon.pdf}} & Please select the most appropriate option. Suppose all the letters are on a Cartesian coordinate system established on the plane, where \textcolor{red}{upward is defined as West}. In which direction does \textcolor{red}{Q lie in relation to P}? \newline   A. East B. West C. South D. North E. Northeast F. Northwest G. Southeast H. Southwest & G. Southeast\\ \hline
        
        Relative Compass Reasoning (number) & \raisebox{-0.5\height}{\includegraphics[width=0.083\textwidth, clip, trim=106.5mm 116mm 163mm 61.6mm]{icon.pdf}} & Please select the most appropriate option. Suppose all the letters are on a a Cartesian coordinate system established on the plane, where \textcolor{red}{upward is defined as East}. In which direction does \textcolor{red}{20 lie in relation to 64}?  \newline  A. East B. West C. South D. North E. Northeast F. Northwest G. Southeast H. Southwest & H. Southwest\\ \hline
    \end{tabular}
\vspace{-2mm}
\end{table*}

To explore direction reasoning, we divide it into spatial and compass directions. Based on human cognitive principles~\cite{haun2011plasticity,levinson2003space}, we consider spatial reasoning—understanding the relative positions of objects (up, down, left, right, and combinations)—as the foundation for compass reasoning, which focuses on geographical directions (north, south, east, west, and combinations). While spatial reasoning is limited to internal image content, compass reasoning requires alignment with real-world direction rules, making it crucial for tasks like navigation and geographical positioning, and key to evaluating a model's ability to apply knowledge in practical scenarios.

Earlier work uses text descriptions for spatial reasoning and path planing~\cite{DBLP:conf/aaai/ShiZL22,DBLP:journals/tmlr/YamadaBLKY24, mirzaee2021spartqa,DBLP:journals/corr/abs-2310-03249,DBLP:conf/aaai/LiH024a}. Inspired by human cognition, wu et al.~\cite{wu2024mind} introduced the Visualization-of-Thought technique to guide LLM through reasoning steps in spatial reasoning. liu et al.~\cite{liu2023visual} and kamath et al.~\cite{kamath2023s} collected a real-world scenario corpus of visual question answering for spatial reasoning. 

However, current spatial reasoning datasets primarily focus on describing the relative positions of objects within images. While models may learn these spatial relationships, there is a lack of benchmarks that evaluate whether models can understand compass directions as they apply to the real world. Compass reasoning is crucial because it reflects the fundamental orientation principles governing the physical world. Without such benchmarks, it is unclear whether models have truly internalized these real-world principles or are merely relying on learned spatial patterns from the data.

\begin{figure}[t]
    \centering
    \includegraphics[scale=0.6, clip, trim=16mm  80mm 135mm 30mm]{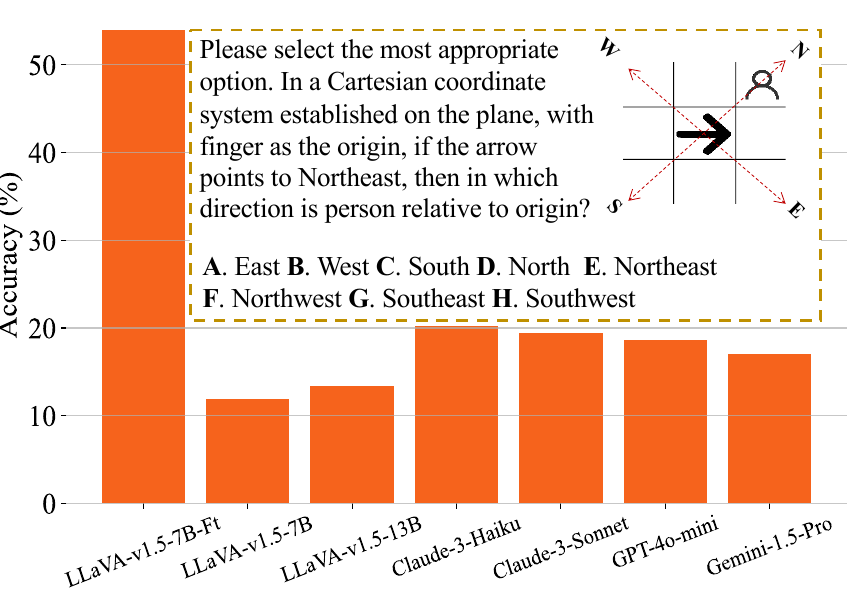}
    \caption{Accuracy of multimodal language models on the Relative Compass Reasoning Task. We fine-tune the LLaVA-7B on CDR and significantly improves to 53.43\%.
}
\vspace{-4mm}
    \label{fig:help}
\end{figure}

In this work, we propose the Compass Direction Reasoning (CDR) benchmark, designed to mainly evaluate the compass reasoning ability of models. 
The CDR dataset contains handcrafted image-question pairs in English. Each pair includes a 2D image and corresponding direction-related questions, which have 8 possible directions and a unique correct answer.
The images in the CDR dataset are symbol-based and consist of three types. The first type is icon images, where the central icon has a clear direction orientation (e.g., a finger, an arrow, etc.) with surrounding objects (e.g., a person, a flower, etc.), as shown in Table~\ref{tab:prompts} image (2). These simple, easily distinguishable icons are designed to emphasize the model's direction reasoning without interference from object categorization. The second type includes letters, randomly placed in the central and surrounding cells without repetition, highlighting spatial relationships, as shown in Table~\ref{tab:prompts} image (3). Similarly, numbers are placed in a grid, as shown in Table~\ref{tab:prompts} image (4).

We evaluate 6 popular MLMs, including LLaVA-7B \& 13B~\cite{liu2024improved}, Claude-3-Haiku \& Sonnet~\cite{anthropic2024claude}, GPT-4o-mini~\cite{OpenAI2024Mini}, and Gemini-1.5pro-flash~\cite{reid2024gemini}. As shown in Figure~\ref{fig:help}, in relative direction reasoning tasks, most models exhibit a high error rate, performing no better than random guessing (12.5\%) and significantly below human-level performance (100\%). Experiments show that further fine-tuning LLaVA-7B on CDR significantly improves its ability to answer direction questions. Our contributions can be summarized as follows: \textit{(i)} We propose CDR, a multimodal direction reasoning benchmark that integrates both spatial and compass reasoning using simple, intuitive, and low-ambiguity images. CDR provides balanced direction distributions, contributing to a more fair and comprehensive evaluation of direction reasoning.
\textit{(ii)} We progressively introduce different types of direction experiments based on the logic of direction reasoning, and we analyze the results across multiple MLMs step by step.
\textit{(iii)} We explore the effects of fine-tuning methods with mixed data and CoT (Chain of Thought) data on LLaVA-7B and provide a detailed analysis of the potential reasons behind their performance.

\vspace{-2mm}
\section{METHODOLOGY}

\subsection{Data Collection}
\paragraph{Image Collection \& Construction}
To prevent complex object categories from influencing the model's direction reasoning, we use simple objects. We begin with a 200$\times$200 canvas (the final image size), and a 3$\times$3 grid is drawn on it. For letter-type images, to establish relationships between the letters, a cell containing a letter (the center letter) is randomly selected, and the surrounding cells are filled with non-repeating letters, allowing for empty cells to avoid repetition. Each letter and its position are recorded to facilitate subsequent questions and automated annotation. Similarly, number-type images follow the same strategy. For icon-type images, we source high-resolution, unambiguous data from the Internet~\cite{iconfont2024}. The central icon is a direction symbol (e.g., arrows, road signs, pointing fingers), and the surrounding objects include human-shaped icons, flowers and so on. We manually annotate the central icon's spatial orientation and record the spatial positions of the surrounding icons relative to it, enabling automatic label generation for question construction.

\paragraph{Question \& Answer Construction}
Based on the collected images, we design prompts with a focus on simple, direct approaches to evaluate fundamental directional reasoning abilities. For evaluation, we use seven types of questioning methods, as shown in Table~\ref{tab:prompts}.

First, we ask about the categories of the symbols, as shown in Table~\ref{tab:prompts} \textit{Object Classification}.  
Second, we question the absolute direction of directional icons, as shown in Table~\ref{tab:prompts} \textit{Spatial Direction Reasoning} and \textit{Compass Direction Reasoning}. This step aims to evaluate whether the model can accurately reason about the absolute direction indicated by the icons, both spatial and compass.

In the third step, we ask about the relative directions of combined icon images, as shown in Table~\ref{tab:prompts} \textit{Relative Spatial Reasoning}. Assuming that the surrounding icons and the central icon are on the same plane, we inquire about their relative spatial positions. Notably, for the letter and number-type data, we only follow the first and third steps, as these types of data do not inherently contain direction information.

We evaluate the model's compass reasoning by specifying an upward direction and asking which compass direction the central icon points to, as shown in Table~\ref{tab:prompts} \textit{Compass Direction Reasoning}.

Finally, as shown in Table~\ref{tab:prompts} \textit{Relative Compass Reasoning}, we use combined icons or letters, specifying the direction of the central element (e.g., an arrow pointing north). Using the central icon or letter as the origin of a coordinate system, we ask about the surrounding elements' relative compass directions. We develop four variations of questioning methods to accurately evaluate the model's generalization and robustness in handling similar tasks. As image (3) in Table~\ref{tab:prompts} shown, the model needs to map the spatial rules to the real world compass direction rules to correspond ``upward" to ``West" to infer the specific orientation of Q relative to P. This requires the model to not only understand relative positions in the image, but also map them to real-world orientation systems.

These strategies generate numerous images and questions with varied spatial relationships, providing diverse and balanced training and testing data for the model.

\subsection{Data Statistics}

CDR includes three types of image data: letter, number and icon-type. By varying the center letter, surrounding letters, and their respective locations, we create a letter dataset consisting of 12,495 training images and 1,275 testing images. For the number dataset, we generate 1,275 testing images. Similarly, by altering the central icon and the positions of the surrounding icons, we produce an icon dataset containing 1,580 icons and 9,144 combined icon images.

The types of questions asked vary depending on the specified direction. For spatial letter direction reasoning, the CDR dataset includes 50,176 training samples and 4,080 testing samples. For compass direction reasoning, it contains 50,176 training samples and 4,080 testing samples for letters, 5,120 testing samples for numbers, and 71,552 training samples with 4,000 testing samples for icons.

\begin{figure}[h!] 
    \centering
    \begin{minipage}[b]{0.24\textwidth}
        \centering
        \includegraphics[scale=0.15, clip, trim=88mm 20mm 105mm 20mm]{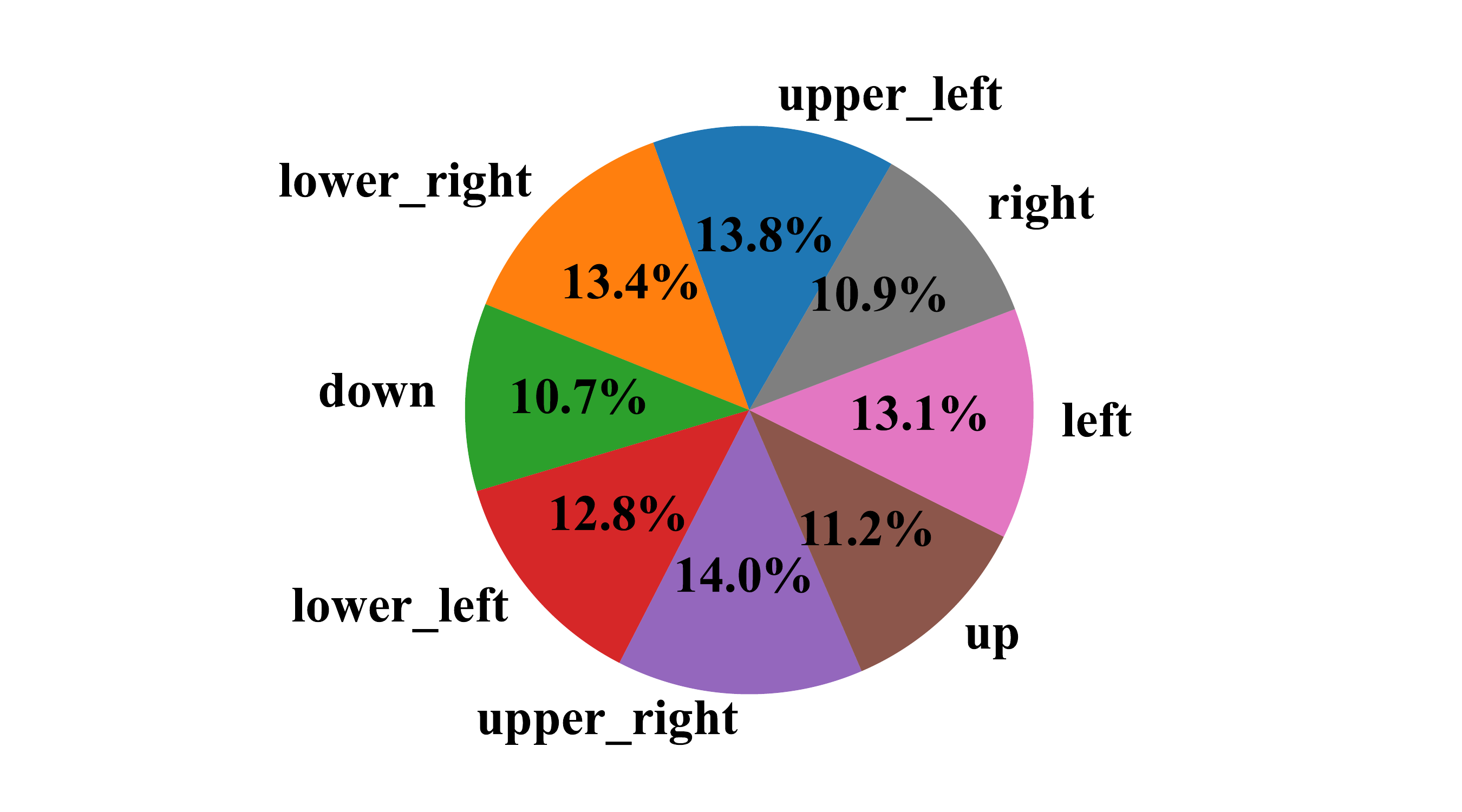}
    \end{minipage}
    \hfill
    \begin{minipage}[b]{0.24\textwidth}
        \centering
        \includegraphics[scale=0.15, clip, trim=78mm 20mm 95mm 20mm]{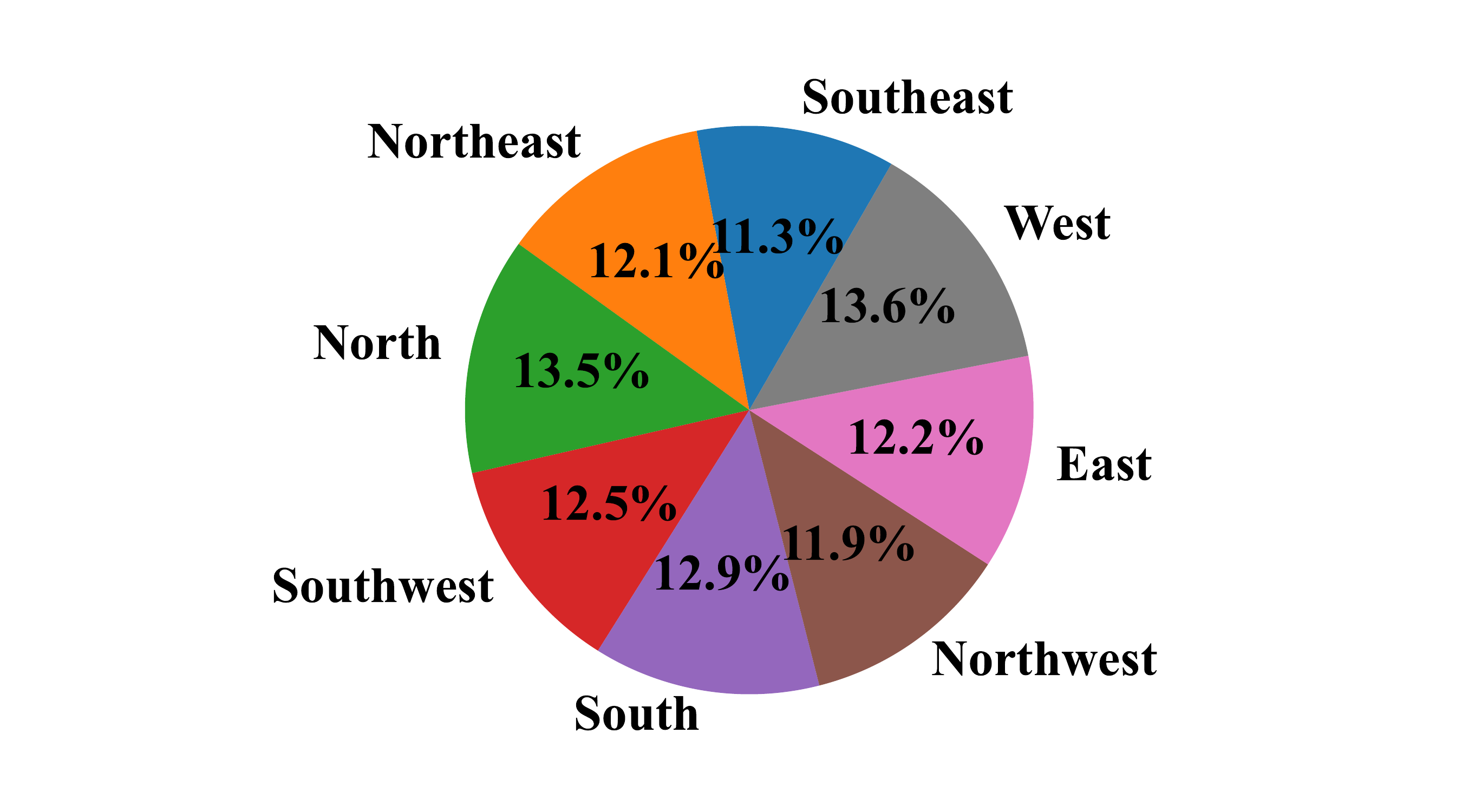}
    \end{minipage}
    \caption{Distribution of CDR answers in spatial and compass reasoning tasks.}
    \label{fig:reasoning_distri}
\end{figure}

The CDR have balanced direction labels for testing both spatial and compass reasoning tasks. As shown in Figure~\ref{fig:reasoning_distri}, this ensures equal representation of each direction category, preventing bias.

\section{EXPERIMENTS AND ANALYSIS}
We evaluate the following models: LLaVA-v1.5-7B, LLaVA-v1.5-13B, Claude-3-Haiku (20240307), Claude-3-Sonnet (20240229), GPT-4o-mini (2024-07-18), and Gemini-1.5pro-flash. The decoding parameters used are: frequency penalty = 0.0, presence penalty = 0.0 and temperature = 0. We set the context window for the LLaVA to 1024 tokens to fine-tune, with other models maintaining their default settings. The experiments are conducted in a zero-shot setting.

\begin{table*}[ht]
    \centering
    \caption{Accuracy of Different MLMs on CDR. Experiment Type Indicates the Specific Task. 
}
    \label{tab:llm result}
    \begin{tabular}{c|c|c|c|c|c|c} % 7个居中对齐的列
        \hline
        \diagbox{Experiment Type}{Models} & LLaVA-7B & LLaVA-13B & Claude-3-Haiku & Claude-3-Sonnet & GPT-4o-mini & Gemini-1.5-Pro \\ \hline
        \textbf{Object Classification (icon)} & 97.72\% & \textbf{99.47}\% & 87.00\% & 70.50\% & 94.91\% & 98.95\% \\ \hline
        \textbf{Object Classification (letter)}& \textbf{100\%} & \textbf{100\%}   & \textbf{100\%}  &\textbf{100\%}  & 99.57\% & \textbf{100\%}   \\ \hline
        \textbf{Object Classification (number)}& 99.89\% & 99.89\%   & \textbf{100\%}  &99.00\%  & 98.89\% & \textbf{100\%}   \\ \hline  \hline
        \textbf{Spatial Direction Reasoning (icon)} &14.08\%  & 25.72\% & 37.00\% & 33.96\% & 61.42\% & \textbf{62.64\%} \\ \hline
        \textbf{Compass Direction Reasoning (icon)} & 12.42\% & 22.73\% &37.14\% & 35.61\% & 51.00\% & \textbf{52.99\%} \\ \hline \hline
        \textbf{Relative Spatial Reasoning (icon)} & 27.27\% & 27.79\% & 16.30\% & 14.90\% & 45.30\% & \textbf{64.80\%} \\ \hline
        \textbf{Relative Spatial Reasoning (letter)} & 15.20\% &22.89\%  & 23.31\% &48.53\%  & 36.35\%  &  \textbf{52.48\%}\\ \hline
        \textbf{Relative Spatial Reasoning (number)}  & 16.99\% &23.61\%  &26.84\%  & \textbf{48.37\%} & 38.83\% & 43.42\% \\ \hline \hline
        \textbf{Relative Compass Reasoning (icon)} & 11.90\% & 13.41\% & 20.26\% & 19.46\% & 18.65\%& 17.04\% \\ \hline
        \textbf{Relative Compass Reasoning (letter)}  & 13.04\% &13.95\%  &14.97\%  & 19.90\% & 15.27\% & \textbf{21.35\%} \\ \hline
        \textbf{Relative Compass Reasoning (number)}  & 13.54\% & 11.82\%  &13.50\%  & \textbf{18.80\%} & 16.78\% & 17.74\% \\ \hline
    \end{tabular}
    \vspace{-2mm}
\end{table*}

\subsection{Object Classification} In the classification task, all models performed well, with LLaVA-7B and LLaVA-13B achieving 97.72\% and 99.47\% accuracy, respectively. The Claude-3 series and Gemini-1.5-Pro also demonstrated high accuracy, showcasing their strong symbol classification capabilities. This further indicates that the models' direction reasoning is not hindered by their ability to recognize the images.

\subsection{Absolute Direction Reasoning}In spatial direction reasoning, model performance is generally poor. LLaVA-7B achieves an accuracy of just 14.08\%, while LLaVA-13B improves slightly to 25.72\%. Claude-3 and Gemini-1.5-Pro perform marginally better in comparison.
For compass direction reasoning, results remain low. LLaVA-7B and LLaVA-13B achieve accuracy rates of 12.42\% and 22.73\%, respectively. In comparison, the Claude-3 series and Gemini-1.5-Pro perform slightly better, but their maximum accuracy of 62.64\% is still far from ideal. This suggests that the models struggle with geographical orientation tasks, likely due to the abstract nature of these concepts, which are beyond the models' current capabilities. The introduction of the compass concept appears to further challenge the models' ability to understand the tasks, resulting in a general decline in their comprehension. This further confirms their limited understanding of real-world direction reasoning. Since letters and numbers do not inherently contain direction information, we only test icon data for these tasks.

\subsection{Relative Direction Reasoning}In the relative direction reasoning task, the model's performance drops significantly for both spatial and compass directions. Specifically, in the relative compass direction reasoning task, LLaVA-7B and LLaVA-13B achieve accuracy rates of only 11.98\% and 11.7\%, respectively, which are much lower than in other tasks. Similarly, other models such as the Claude-3 series and GPT-4o-mini also perform poorly, with accuracy rates mostly between 10\% and 20\%. This significant decline in accuracy suggests that the models struggle with the increased complexity involved in relative orientation reasoning.
One possible explanation for these results is that relative direction reasoning requires a higher level of abstraction and the ability to process multiple orientation cues simultaneously. The models may lack the ability to integrate these cues effectively, leading to confusion when determining the correct relative orientation. Additionally, the inherent complexity of relative orientation tasks, which involve understanding the relationship between multiple objects and directions, may exceed the models' current capabilities, particularly if the training data did not sufficiently cover such scenarios. This highlights the need for more advanced training strategies and datasets that better capture the intricacies of relative direction reasoning.

\begin{table*}[h]
    \centering
    \caption{The Fine-tuning Results of LLaVA-7B on Relative Compass Reasoning (icon) task.}
    \label{tab:ft_results}
    \begin{tabular}{c|c|c|c|c|c|c|c|c} % 去除左右边框
        \hline
         & \multicolumn{8}{c}{\textbf{Fine-tuning Dataset}} \\ \hline
     \textbf{7B\_Base}&\textbf{All} & \textbf{All+20K}&\textbf{All+40K} &\textbf{All+60K} &\textbf{All+80K} &\textbf{All+100K}  & \textbf{CoTAll}&\textbf{CoTAll+40K} \\ \hline
        11.90\% & 10.69\% & 20.87\% & 31.05\% & 24.50\% & 15.93\%& 19.45\% &34.78\% & \textbf{53.43\%}\\ \hline
    \end{tabular}
    \vspace{-2mm}
\end{table*}

\subsection{Fine-tuning Results}

We fine-tune the LLaVA-7B using the Relative Compass Reasoning (icon) training set from CDR, along with randomly mixed data from \textit{llava\_v1\_5\_mix665k.json}. The results from various dataset combinations are shown in Table~\ref{tab:ft_results}.
\textbf{All} refers to fine-tuning with the full CDR training dataset, while \textbf{20K, 30K, 40K}, etc., refer to fine-tuning with randomly selected subsets of 20K, 30K, and 40K samples, respectively.

Using the all Relative Compass Reasoning training data in (71552 samples), the model only get 11.90\% accuracy on Relative Compass Reasoning task. It suggests that the model struggles to effectively learn the underlying rules of orientation purely by associating options with direction questions.
\vspace{-2mm}
\begin{table}[ht]
\centering
\caption{An example of CoT Instruction fine-tuning data in CDR.}
\label{tab:cot}
\begin{tabular}{>{\raggedright\arraybackslash}p{8.3cm}}
\hline
\textless \textbf{Image}\textgreater\\
\textbf{Question:} Please select the most appropriate option. In a Cartesian coordinate system established on the plane, with finger as the origin, if index finger is points to Northeast, then in which direction is person relative to origin? \\
A. East B. West C. South D. North E. Northeast F. Northwest G. Southeast H. Southwest\\
\hline \hline
\textbf{Answer}: H.Southwest \\
\hline
\textbf{CoT Answer}: Take the finger as the origin, the finger points to the down of the image and is in the Northeast direction.
Therefore, the left side of the image is the SouthEast, \textcolor{red}{the top side is the Southwest}, the bottom side of the image is the Northeast, the right side of the image is the Northwest, the upper left side of the image is the South, the upper right side of the image is the West, the lower left side of the image is the East, the lower right side of the image is the North.
Since the person is located \textcolor{red}{in the top of the image}, it can be inferred that the person is located Southwest of the origin, so select H. Southwest. \\ \hline
\end{tabular}

\end{table}

Based on the \textit{All} dataset, we progressively added out-of-domain datasets for mixed training. From \textit{All+20K} to \textit{All+100K}, we observe that \textit{All+40K} achieved the highest accuracy at 31.05\%. This may be due to the optimal mix ratio (approximately 2:1), which provided better generalization for the model. At the same time, this indicates that simply increasing the proportion of mixed data may not be the key to improving performance in relative compass reasoning.

Furthermore, we create CoT data for the relative compass reasoning task, enriching responses with step-by-step compass direction reasoning, as shown in the \textit{CoT Answer} in Table~\ref{tab:cot}. Using rule-based logical reasoning, the model's accuracy improves significantly to 34.78\% (\textit{CoTAll}), demonstrating the effectiveness of teaching the model the relationship between spatial direction reasoning and compass direction reasoning. After learning from this CoT data, the model first describes the central object and its orientation, then discusses compass directions for each spatial direction, and finally determines the compass directions of surrounding objects relative to the origin. By learning step-by-step reasoning, the model better understands and reasons through complex direction relationships.

We select the best mix ratio from previous mixed data experiments (\textit{All+40K}) and replace the \textit{All} data with CoT data (\textit{CoTAll+40K}). The model's performance improves, suggesting that CoT data provides clear, step-by-step reasoning, helping the model focus on key task elements. Mixed data adds complexity and variation, prompting the model to refine its direction inference. This combination enhances learning efficiency, allowing the model to better master direction inference rules. Additionally, adding ``\textit{think step by step}'' to the prompt without fine-tuning shows LLaVA-7B ignoring CoT instructions and directly outputting answers, resulting in poorer performance.

Experiments show that while models can recognize objects in images, most struggle with accurate direction reasoning. Advanced models like GPT-4o-mini and Gemini-1.5-Pro perform well in spatial reasoning but drop sharply in compass reasoning. In relative compass tasks, most models perform near random guessing, highlighting limitations in handling complex multi-object reasoning.
Our fine-tuning experiments on LLaVA-7B reveal difficulties in learning direction relationships from single option and answer. Mixed data improves generalization, while CoT fine-tuning helps the model understand spatial and compass relationships, enhancing its real-world direction reasoning capabilities.

Although mixdata and CoT fine-tuning improve the model's CDR capabilities, its accuracy still falls far short of human-level performance. Enhancing the model's CDR abilities and enabling it to truly learn direction cognition remains a key focus of our future work.

\section{CONCLUSIONS}
In this paper, we explore the ability of MLMs in compass direction reasoning, as existing research predominantly focuses on spatial reasoning. To address this gap, we propose CDR, the first multimodal benchmark specifically designed for compass direction reasoning. CDR provides a comprehensive evaluation framework for both spatial and compass tasks, offering balanced directional distributions for precise and thorough evaluation with over 100K training and testing samples. Its simple, low-ambiguity design ensures fair assessment across diverse directional challenges. Fine-tuning LLaVA-7B on CDR demonstrates significant improvements, particularly in compass-based tasks. 
In future work, we plan to expand the dataset by incorporating more complex, real-world images (such as map data) and exploring 3D directional reasoning to further enhance the evaluation of models' directional reasoning abilities.

\vspace{1em}

\newpage
\newpage
\bibliography{ref}
\bibliographystyle{IEEEtran}
\end{document}